\documentclass[conference]{IEEEtran}
\usepackage[pdftex]{graphicx} % Required for including pictures
\usepackage[pdftex,linkcolor=black,pdfborder={0 0 0}]{hyperref} % Format links for pdf
\usepackage{cite}
\usepackage{multirow}
\usepackage{amsmath}
\usepackage{hyperref}
\usepackage{algorithmic}
\usepackage{array}
\usepackage{dblfloatfix}
\usepackage{url}
\usepackage[table]{xcolor}

\begin{document}
\IEEEoverridecommandlockouts
\IEEEpubid{\makebox[\columnwidth]{\textcolor{blue}{Published in IEEE ISM 2021. Please cite \cite{9666132}
when referencing this paper.}
 \hfill} \hspace{\columnsep}\makebox[\columnwidth]{ }}
\title{Comprehensive Saliency Fusion for Object Co-segmentation}
\author{
\IEEEauthorblockN{
Harshit Singh Chhabra}
\IEEEauthorblockA{
Indraprastha Institute of Information Technology Delhi}
\and
\IEEEauthorblockN{Koteswar Rao Jerripothula}
\IEEEauthorblockA{
Indraprastha Institute of Information Technology Delhi}
}
\maketitle

\begin{abstract}
Object co-segmentation has drawn significant attention in recent years, thanks to its clarity on the expected foreground, the shared object in a group of images. Saliency fusion has been one of the promising ways to carry it out. However, prior works either fuse saliency maps of the same image or saliency maps of different images to extract the expected foregrounds. Also, they rely on hand-crafted saliency extraction and correspondence processes in most cases. This paper revisits the problem and proposes fusing saliency maps of both the same image and different images. It also leverages advances in deep learning for the saliency extraction and correspondence processes. Hence, we call it comprehensive saliency fusion. Our experiments reveal that our approach achieves much-improved object co-segmentation results compared to prior works on important benchmark datasets such as iCoseg, MSRC, and Internet Images.
\end{abstract}


\section{Introduction}
Object co-segmentation is the task of extracting common objects as foregrounds in a group of images. Compared to the single image segmentation, where it's difficult to ascertain what's foreground, we have a clear-cut definition of what foreground we wish to extract in the object co-segmentation task. Rother $et. al.$ \cite{rother2006cosegmentation} first introduced the concept of co-segmentation by developing a histogram matching method to extract common parts from a pair of images. However, Vicente $et. al.$ \cite{vicente2011object} were the first to propose that co-segmentation should be about things (or objects) and there is a need to incorporate a measure of objectness in the models. They proposed a Random Forest classifier to find the similarity between a pair of proposed segmentation of objects in each image of a group followed by the A* search algorithm \cite{andres2010empirical} to find the segmented objects with maximum similarity score. The major applications of object co-segmentation include image grouping\cite{kim2012hierarchical}, object recognition\cite{gallagher2008clothing}, and object tracking\cite{tsai2016semantic} which are among the fundamental tasks of Computer Vision .\par
% \begin{figure}
% \centering
% \includegraphics[width=0.48\textwidth, height = 2.5cm]{Img1.PNG}
% \vspace{-2.5mm}
% \caption{Co-Segmentation over an image pair}
% \label{fig_1}
% \vspace{-7mm}
% \end{figure}

The problem of object co-segmentation is highly correlated to finding the common regions of interest in a group of images. \cite{jerripothula2014automatic, jerripothula2015group} approach the problem of co-segmentation by trying to highlight these common regions of interest. \cite{meng2012object} proposed a shortest path algorithm between the salient regions of images, \cite{rubinstein2013unsupervised} found saliency maps and the computed dense correspondence for all the similar images for obtaining matching scores for pixels across the images. Geometric Mean Saliency (GMS) \cite{jerripothula2014automatic} uses dense SIFT Correspondence \cite{liu2010sift} for each pair of images to align saliency maps and fuses them. \cite{jerripothula2015group} extended \cite{jerripothula2014automatic} by making it more efficient through introducing the key image concept. \cite{tao2017image} use saliency prior and superpixels to partition images into foreground and background. They calculate several feature level descriptors on superpixel level and organized in Bag of Words (BoW) to perform clustering of common foregrounds. All of these use single saliency source. Rather than going through the same trend, we propose the usage of multiple saliency sources to benefit from all of them. \\

\indent In the last decade, there has been a immensely growing interest towards deep learning in the computer vision coomunity, thanks to better availability of computational resources and platforms for large-scale annotations. A large variety of deep neural networks have been proposed for the task of object co-segmentation. \cite{li2018deep, chen2018semantic} use deep siamese networks to perform object co-segmentation. \cite{li2018deep} use a pre-trained VGG16 network for feature extraction and pass them to a correlation layer to find the similarities among a pair of images as inspired by Flownet \cite{dosovitskiy2015flownet}. The features from the encoder are combined with the correspondence obtained from the correlation layer and passed to the siamese decoder to obtain the output co-segmentation masks. \cite{chen2018semantic} developed a similar architecture to perform deep object co-segmentation by passing the global image features obtained from pre-trained VGG16 model to a channel-wise attention model which trains a fully connected network. The output features are then upsampled and sent to a convolution network which acts as a siamese decoder and generates the cosegmentation mask for the input images. Recently, \cite{zhang2020deep} developed a unique deep learning architecture which uses HRNet\cite{sun2019deep} pre-trained on ImageNet as a backbone to extract image features and then capture the correlation between features using a spatial modulator to generate a mask which can localize the common foreground object. They also developed a semantic modulator which acts as a supervised image classification model. The combined outputs of both modulators is able to co-segment common objects by using the multi-resolution image features. \\

\indent However, it is well-known that deep learning based co-segmentation methods require training which can make these approaches class specific. The performance of these models on unseen data will largely depend on the similarity between the seen and unseen classes. In such a scenario, saliency based co-segmentation methods might give superior performance because they are based on generic saliency methods. In the proposed approach, we exploit the inherent genericness of deep learning based saliency methods to identify common salient objects. An important aspect in a co-segmentation method is finding the correspondence between the images in a group. Usually, it's done through hand-crafted feature descriptors like SIFT \cite{liu2010sift}, GSS \cite{deselaers2010global}, etc. However these traditional methods do not work well when there are large viewpoint or illumination changes. In the proposed approach we use deep ResNet features and pre-trained DGC Net\cite{melekhov2019dgc} to establish global and local correspondences, respectively.\\

\indent We make two contributions: (1) using multiple saliency sources and (2) using learning based saliency extraction and correspondence processes for effective object co-segmentation. 

%\indent The basic idea for our proposed approach is to perform clustering of images based on deep features extracted from the pre-trained ResNet50 \cite{he2016deep} model and select a representative image for each subgroup. We create multiple deep saliency maps for each image in the dataset and then align the saliency maps of member images of each subgroup to the representative by using a pre-trained Dense Correspondence Network and fuse these warped maps to make a co-saliency map for the representative. This co-saliency map is transmitted back to individual members to make co-saliency maps for each image. Finally, we use the Grabcut algorithm to convert these co-saliency maps to output co-segmentation masks. The proposed approach is able to achieve competitive results on standard co-segmentations datasets like iCoseg\cite{batra2010icoseg}, MSRC\cite{shotton2006textonboost}, and Internet\cite{rubinstein2013unsupervised}.
\begin{figure}
\centering
\includegraphics[width=1\linewidth]{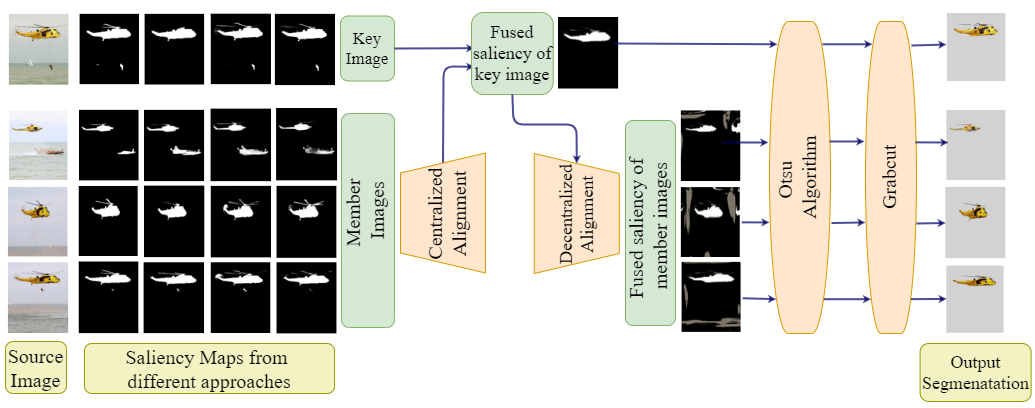}
\vspace{-9mm}
\caption{Proposed Approach}
\label{fig:fig_2}
\end{figure}

\section{Methodology}

\textbf{Notations and Overview:} Let's consider a group of $m$ images $\mathcal{I}=\{I_1, I_2, …I_m\}$, which is divided into $K$ sub-groups, following \cite{jerripothula2014automatic}. If $V_i$ denotes the sub-group label of $I_i$, we can define $k^{th}$ sub-group of images as $G_k=\{I_i|V_i=k\}$, where $k\in\{1,\cdots,K\}$. Let's denote the cluster representative of $k^{th}$ cluster with the notation $g_k$. 

As far as saliency maps are concerned, let's denote $j^{th}$ saliency map of $i^{th}$ image as $S^j_i$, where $j\in \{1,\cdots,L\}$, if we consider $L$ saliency sources. Let the warping function between two images, say $I_u\rightarrow I_v$, be denoted as $W^{I^v}_{I_u}$.   

In Fig.~\ref{fig:fig_2}, we show how different fused saliency maps are generated in a sub-group and eventual object co-segmentation masks are obtained. The core idea is to fuse the saliency maps available in a sub-group through appropriate alignments, which are of two types: centralized and decentralized. The centralized and decentralized alignments lead to fused saliency maps for the key image and member images, respectively. Let's discuss all this in detail.     

\textbf{ Sub-group Formation:}
In a given group of images, there may be lot of shape and pose variance across the images, which is not ideal for saliency fusion. Also, we need not process all of them together. We must note that an image needs only a couple of other neighboring images to carry out object co-segmentation. Hence, we divide the group into $K$ sub-groups via k-means clustering on global features extracted using ResNet50 pre-trained model (on ImageNet dataset\cite{deng2009imagenet}). We perform silhouette analysis \cite{rousseeuw1987silhouettes} to find the best $K$. We also identify the images nearest to the cluster centers as key images.

\textbf{Saliency Map Generation:}
Current saliency fusion-based co-segmentation approaches mostly utilize a single saliency source. In contrast, we extract four saliency maps (which means $J=4$) using different deep saliency detection methods, namely PoolNet \cite{liu2019simple}, EGNet \cite{zhao2019egnet}, BASNet \cite{qin2019basnet}, U2Net \cite{qin2020u2}. The main motivation behind using multiple methods is to benefit from all the sources, instead of just one.

\textbf{Saliency Fusion:}
Our proposed framework has two stages: (i) centralized alignment, and (ii) decentralized alignment.  
 
For generating fused saliency map of the key image, we align all the available saliency maps in the sub group to the key image. We call this centralized alignment. For making alignments, we use the pre-trained DGC-Net \cite{melekhov2019dgc} model throughout our work. Such learning-based models provide us excellent correspondences between the pixels from two different images, which can be used to align saliency maps as well, not just images. First let's collect all the candidate saliency maps for the key-image in a sub-group after proper alignment, as given below:  

\begin{equation}
    C_{k}= \{W^{g_k}_{I_i}(S_i^j)|I_i\in{G_k},j\in\{1,\cdots,L\}\}
\end{equation}
where $C_k$ denotes the collection. There will be a total of $L*|G_k|$ candidate saliency maps in the sub-group $G_k$ for the key image $g_k$. Here $|\cdot|$ denotes cardinality. Since all these candidate saliency maps will be of same size, because of aligning to one image, the key-image, we can denote the collection of candidate saliency values of a pixel $p$ as $C_k^p$. We fuse all these candidate values to generate fused saliency map $F_{g_k}$ for the key image $g_k$ in the following way:
\begin{equation}
    F_{g_k}(p)=median(C_k^p)
\end{equation}
where we are basically finding the median of all the candidate saliency values for any pixel $p$. Having obtained the fused saliency map for the key-images, to obtain such map for sub-group members, we perform what we call decentralized alignment, i.e., aligning fused saliency map of key image to the member images. Note here that, in the centralized alignment discussed earlier, the alignment was opposite, i.e. aligning saliency maps of member images to the key image. Once we know the key-image to member image correspondences, we can align the fused saliency obtained for the key image to the member images. For instance, we can obtain fused saliency map for a member image $I_i$ in the following way:
\begin{equation}
    F_{I_i}=W^{I_i}_{g_{V_i}}(F_{g_{V_i}})
\end{equation}
where $F_{I_i}$ denotes the fused saliency map for image $I_i$.  

\textbf{Object Co-segmentation Masks:}
Having obtained fused saliency map for every image, we now need to generate their object co-segmentation masks. For that, we first apply the Otsu algorithm on those maps to generate foreground and background seeds. These seeds are then passed along with the image to Grabcut \cite{rother2004grabcut} algorithm to extract our expected foreground, the shared object.

\section{Experimental Results}
We conduct all our experiments on three standard benchmark datasets of object co-segmentation research, namely MSRC, iCoseg and Internet Images. The most common evaluation metrics used for this task in literature are Jaccard Similarity Score ($\mathcal{J}$) and Precision Score ($\mathcal{P}$). We use both of them. Jaccard Similarity is defined as the intersection over union (IoU) of the resultant co-segmentation mask and groundtruth mask whereas Precision is used to measure the percentage of correctly labeled foreground and background pixels.\\

\begin{figure}[h]
\centering
\includegraphics[width=1\linewidth]{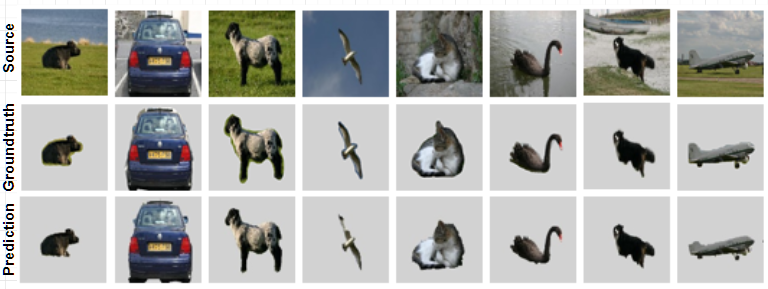}
\vspace{-8mm}
\caption{Sample qualitative results of our approach on MSRC Dataset}
\label{fig_4}

\includegraphics[width=1\linewidth]{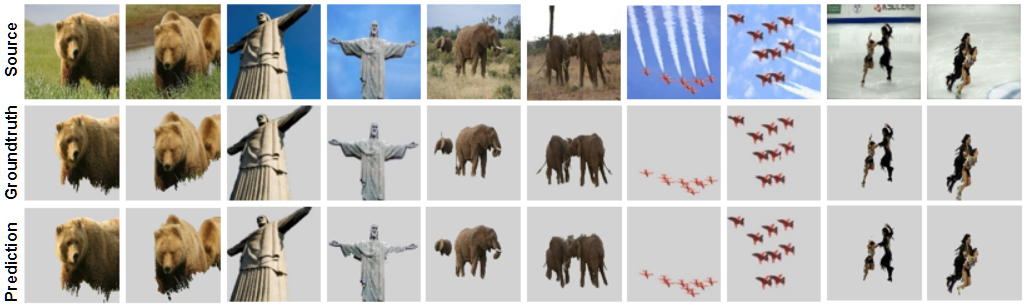}
\vspace{-8mm}
\caption{Sample qualitative results of our approach on iCoseg dataset}
%\caption{}
%\hfil
\label{fig_5}

%\vspace{-1mm}

\includegraphics[width=1\linewidth]{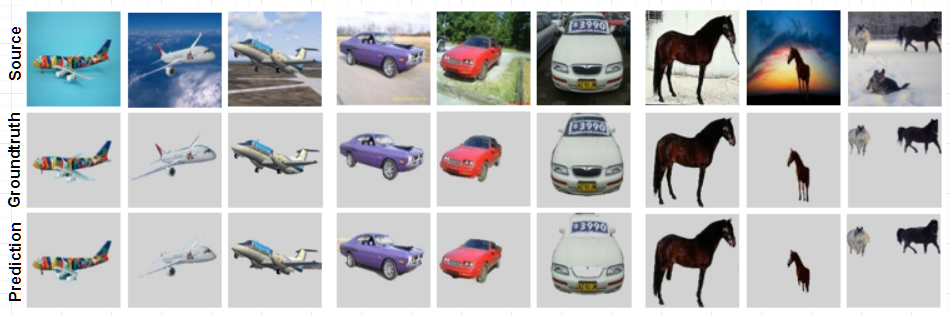}
\vspace{-8mm}
\caption{Sample qualitative results of our approach on Internet Images dataset}    
\label{fig_6}

\includegraphics[width=1\linewidth]{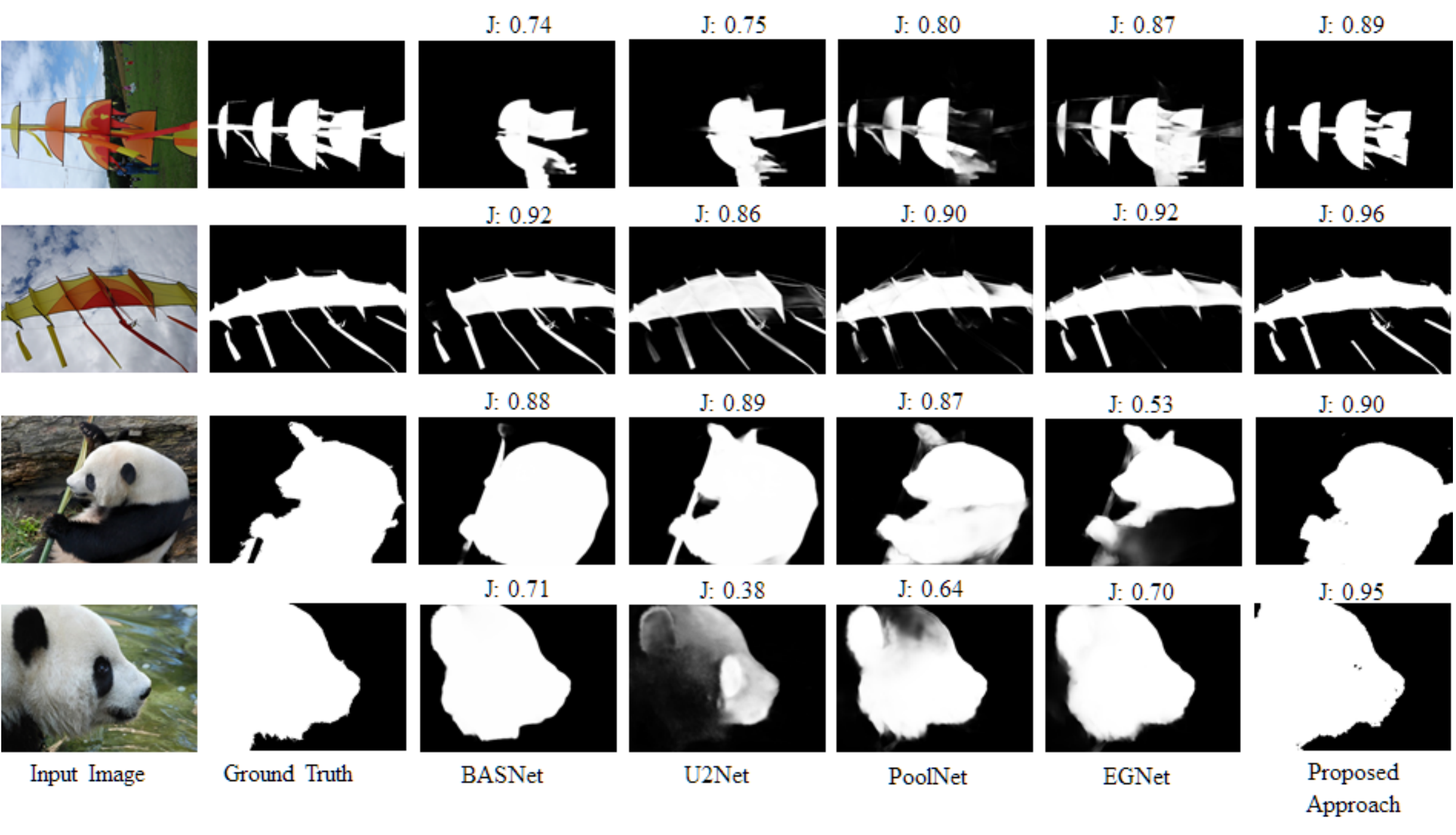}
\vspace{-9mm}
\caption{Comparisons with segmentation scores of individual saliency methods through Otsu segmentation followed by GrabCut, without any type of fusion.}
\label{fig_3}

\end{figure}

We use the standard subset \cite{rubinstein2013unsupervised} of MSRC with classes cat, bird, dog, sheep, car, plane and cow, where each class contains 10 images. It can be seen from Table~\ref{Table 1} that our proposed approach obtains the best $\mathcal{J}$ score and the second best $\mathcal{P}$ score. We provide our sample qualitative results on this dataset in Fig.~\ref{fig_4}.

\begin{table}[h!]
%% increase table row spacing, adjust to taste
\renewcommand{\arraystretch}{1.5}
\vspace{-2mm}
% if using array.sty, it might be a good idea to tweak the value of
% \extrarowheight as needed to properly center the text within the cells
\caption{Comparison of our results with that of state of the art methods on MSRC and iCoseg datasets. Numbers in red and blue indicate the best and the second-best values.}
\label{Table 1}
\vspace{-3mm}
%% Some packages, such as MDW tools, offer better commands for making tables
%% than the plain LaTeX2e tabular which is used here.
\centering
\begin{tabular}{|c|c|c|c|c|}
    \hline
\textbf{Methodology}    & \multicolumn{2}{c|}{\textbf{MSRC Dataset}}& \multicolumn{2}{c|}{\textbf{iCoseg Dataset}}  \\
\hline
   &$\mathcal{P}$  &   $\mathcal{J}$&$\mathcal{P}$&$\mathcal{J}$\\
    \hline
Rubenstein 2013\cite{rubinstein2013unsupervised}   &   \textcolor{red}{92.2}  &   0.75  &   89.6  &   0.68 \\
    \hline
Faktor 2013\cite{faktor2013co}   & 92.0   &    0.77 &  -   &    0.78\\
    \hline
Jerripothula 2016\cite{jerripothula2016image}   & 88.7   &    0.71   & 91.9   &    0.72   \\
    \hline
Ren 2018\cite{ren2018mutual} & - & 0.72 & - & 0.74\\
\hline
Jerripothula 2018\cite{jerripothula2018quality} & 89.7 & 0.74 & 91.8 & 0.72  \\
\hline
Tsai 2019\cite{tsai2019image} & 86.5 & 0.68 & 90.8 & 0.72\\
\hline
Tao 2019\cite{tao2019multi} & 89.8 & 0.72 & \textcolor{blue}{93.2} & 0.76\\
\hline
Jerripothula 2021\cite{jerripothula2021image} & - & \textcolor{blue}{0.79} & - & \textcolor{blue}{0.81}\\
\hline
Proposed\ Approach & \textcolor{blue}{92.1} & \textcolor{red}{0.84} & \textcolor{red}{94.4} & \textcolor{red}{0.88}\\
\hline
\end{tabular}
\vspace{-4mm}
\end{table}
\begin{table}[h!]
%% increase table row spacing, adjust to taste
\renewcommand{\arraystretch}{1.3}
% if using array.sty, it might be a good idea to tweak the value of
% \extrarowheight as needed to properly center the text within the cells
\caption{Comparison of our results with that of state of the art methods on Internet Images dataset. Numbers in red and blue indicate the best and the second-best values.}
\label{Table 3}
%% Some packages, such as MDW tools, offer better commands for making tables
%% than the plain LaTeX2e tabular which is used here.
\vspace{-3mm}
\centering
\setlength\tabcolsep{3 pt}
\begin{tabular}{|c|c|c|c|c|c|c|c|c| %|p{3cm}|p{1cm}|p{1cm}|p{1cm}|p{1cm}|p{1cm}|p{1cm}|p{1cm}|p{1cm}|
}
    \hline
\textbf{Methodology}    & \multicolumn{2}{c|}{\textbf{Airplane}}& \multicolumn{2}{c|}{\textbf{Car}}& \multicolumn{2}{c|}{\textbf{Horse}}& \multicolumn{2}{c|}{\textbf{Average}}\\
    \hline
   &   $\mathcal{P}$  &   $\mathcal{J}$  &   $\mathcal{P}$  &   $\mathcal{J}$  &   $\mathcal{P}$  &   $\mathcal{J}$  &   $\mathcal{P}$  &   $\mathcal{J}$  \\
    \hline
Rubenstein 2013\cite{rubinstein2013unsupervised} & 88.0 & 0.56 & 85.4 & 0.64 & 82.8 & 0.52 & 82.7 & 0.43\\
    \hline
Jerripothula 2016\cite{jerripothula2016image} & 90.5 & 0.61 & 88.0 & 0.71 & 88.3 & 0.61 & 88.9 & 0.64\\
    \hline
Quan 2016\cite{quan2016object} & 91.0 & 0.56 & 88.5 & 0.67 & 89.3 & 0.58 & 89.6 & 0.60\\
\hline
Tao 2017\cite{tao2017image} & 79.8 & 0.43 & 84.8 & 0.66 & 85.7 & 0.55 & 83.4 & 0.55\\
\hline
Ren 2018\cite{ren2018mutual} & 88.3 & 0.48 & 83.5 & 0.62 & 83.2 & 0.49 & 85.0 & 0.53\\
\hline
Tao 2019\cite{tao2019multi} & 92.4 & 0.63 & 91.9 & 0.78 & \textcolor{blue}{90.1} & 0.62 & 91.4 & 0.68\\
\hline
Zhang 2020\cite{zhang2020deep} & \textcolor{red}{94.8} & \textcolor{blue}{0.70} & \textcolor{blue}{91.6} & \textcolor{red}{0.82} & \textcolor{red}{94.4} & \textcolor{blue}{0.70} & \textcolor{red}{93.6} & \textcolor{blue}{0.74}\\
\hline
Proposed\ Approach & \textcolor{blue}{93.4} & \textcolor{red}{0.85} & \textcolor{red}{92.1} & \textcolor{blue}{0.81} & 89.1 & \textcolor{red}{0.77} & \textcolor{blue}{91.5} & \textcolor{red}{0.81} \\
\hline
\end{tabular}
\end{table}

%\vspace{8mm}

The iCoseg dataset \cite{batra2010icoseg} was proposed initially for the task of interactive co-segmentation but is often used to analyze the performance of automatic co-segmentation approaches as well. It contains 643 images belonging to 38 classes and is sufficiently challenging due to multiple objects and varying camera angles. It can be seen from Table~\ref{Table 1} that our proposed approach gives the best $\mathcal{J}$ and $\mathcal{P}$ scores among all the methods. We provide a few sample qualitative results of our method on this dataset in Fig.~\ref{fig_5}.
%\cite{zhang2020deep} was not considered for comparison in Table~\ref{Table 2} as their results are calculated on a subset of iCoseg dataset.\\

\indent Internet Images dataset \cite{rubinstein2013unsupervised} is a very challenging dataset because it comprises of a few noise images as well, meaning they do not contain the shared object. The standard subset of this dataset used for object co-segmentation contains 100 images for each class. Table~\ref{Table 3} gives a detailed analysis of class-wise and average performance for various methods on this subset of Internet Images dataset. It can be seen that our approach obtains the best average $\mathcal{J}$ and second best average $\mathcal{P}$ scores. Our method obtained at least second rank in almost all the cases except $\mathcal{P}$ of Horse category. We provide a few sample qualitative results of our method on this dataset in Fig.~\ref{fig_6}.

\begin{table}[h!]
%% increase table row spacing, adjust to taste
\renewcommand{\arraystretch}{1.5}
% if using array.sty, it might be a good idea to tweak the value of
% \extrarowheight as needed to properly center the text within the cells
\caption{Comparison between single saliency approach and our multiple saliency fusion approach. Numbers in red and blue indicate the best and the second-best values.}
\label{Table 4}
\vspace{-3mm}
%% Some packages, such as MDW tools, offer better commands for making tables
%% than the plain LaTeX2e tabular which is used here.
\centering
\begin{tabular}{|p{1cm}|p{1.5cm}|p{0.45cm}|p{0.45cm}|p{0.45cm}|p{0.45cm}|p{0.45cm}|p{0.45cm}|}
    \hline
\multicolumn{2}{|c|}{\multirow{2}{*}{\textbf{Approach}}}    & \multicolumn{2}{c|}{\textbf{MSRC} \cite{shotton2006textonboost}}& \multicolumn{2}{c|}{\textbf{Internet} \cite{rubinstein2013unsupervised}}& \multicolumn{2}{c|}{\textbf{iCoseg}\cite{batra2010icoseg}}\\\cline{3-8}
 \multicolumn{2}{|c|}{}   & $\mathcal{P}$  &   $\mathcal{J}$  &   $\mathcal{P}$  &   $\mathcal{J}$  &   $\mathcal{P}$  &   $\mathcal{J}$  \\\hline
\multirow{4}{1em}{Single Saliency} & BASNet \cite{qin2019basnet} & 89.7 & 0.79 & 89.8 & 0.79 & 93.2 & \textcolor{blue}{0.86}\\\cline{2-8}
 & U2Net \cite{qin2020u2} & 90.1 & 0.79 & 90.0 & \textcolor{blue}{0.80} & 92.7 & 0.85\\\cline{2-8}
 & PoolNet \cite{liu2019simple} & \textcolor{blue}{90.6} & \textcolor{blue}{0.80} & \textcolor{blue}{91.1} & \textcolor{blue}{0.80} & 93.0 & 0.85\\\cline{2-8}
 & EGNet \cite{zhao2019egnet} & 90.4 & \textcolor{blue}{0.80} & 90.5 & \textcolor{blue}{0.80} & \textcolor{blue}{93.3} & \textcolor{blue}{0.86}\\
\hline
Multiple Saliency & Proposed\ Approach & \textcolor{red}{92.1} & \textcolor{red}{0.84} & \textcolor{red}{91.5} & \textcolor{red}{0.81} & \textcolor{red}{94.4} & \textcolor{red}{0.88}\\
\hline
\end{tabular}
\end{table}

\textbf{Discussion:} In Table~\ref{Table 4}, we show the efficacy of using multiple saliency sources by comparing with the cases when there is only single saliency source available, i.e. $L=1$. It can be seen that our proposed approach performs better than all the cases involving just one saliency source. It clearly demonstrates that using multiple saliency sources is indeed beneficial. In Fig.~\ref{fig_3}, we also demonstrate how the proposed method performs better than individual saliency methods, if we were to simply apply Otsu algorithm followed by GrabCut on them, without performing any type of fusion.

\section{Conclusion}
We propose a comprehensive saliency fusion framework for performing object co-segmentation. In this framework, unlike previous saliency fusion-based methods, we rely on multiple saliency sources and use deep learning based saliency extraction and fusion processes. Our experimental analysis indicates that we are able to perform better than several state-of-the-art methods. 

\section*{Acknowledgements}
This work was supported by Infosys Centre for Artificial Intelligence, IIIT-Delhi, and  IIIT-Delhi PDA Grant.

\bibliographystyle{IEEEtran}
\bibliography{main}

\begin{thebibliography}{10}
\providecommand{\url}[1]{#1}
\csname url@samestyle\endcsname
\providecommand{\newblock}{\relax}
\providecommand{\bibinfo}[2]{#2}
\providecommand{\BIBentrySTDinterwordspacing}{\spaceskip=0pt\relax}
\providecommand{\BIBentryALTinterwordstretchfactor}{4}
\providecommand{\BIBentryALTinterwordspacing}{\spaceskip=\fontdimen2\font plus
\BIBentryALTinterwordstretchfactor\fontdimen3\font minus
  \fontdimen4\font\relax}
\providecommand{\BIBforeignlanguage}[2]{{%
\expandafter\ifx\csname l@#1\endcsname\relax
\typeout{** WARNING: IEEEtran.bst: No hyphenation pattern has been}%
\typeout{** loaded for the language `#1'. Using the pattern for}%
\typeout{** the default language instead.}%
\else
\language=\csname l@#1\endcsname
\fi
#2}}
\providecommand{\BIBdecl}{\relax}
\BIBdecl

\bibitem{rother2006cosegmentation}
C.~Rother, T.~Minka, A.~Blake, and V.~Kolmogorov, ``Cosegmentation of image
  pairs by histogram matching-incorporating a global constraint into mrfs,'' in
  \emph{CVPR 2006}, vol.~1.\hskip 1em plus 0.5em minus 0.4em\relax IEEE, pp.
  993--1000.

\bibitem{vicente2011object}
S.~Vicente, C.~Rother, and V.~Kolmogorov, ``Object cosegmentation,'' in
  \emph{CVPR 2011}.\hskip 1em plus 0.5em minus 0.4em\relax IEEE, 2011, pp.
  2217--2224.

\bibitem{andres2010empirical}
B.~Andres, J.~H. Kappes, U.~K{\"o}the, C.~Schn{\"o}rr, and F.~A. Hamprecht,
  ``An empirical comparison of inference algorithms for graphical models with
  higher order factors using opengm,'' in \emph{Joint Pattern Recognition
  Symposium}.\hskip 1em plus 0.5em minus 0.4em\relax Springer, 2010, pp.
  353--362.

\bibitem{kim2012hierarchical}
E.~Kim, H.~Li, and X.~Huang, ``A hierarchical image clustering cosegmentation
  framework,'' in \emph{2012 IEEE CVPR}.\hskip 1em plus 0.5em minus 0.4em\relax
  Ieee, 2012, pp. 686--693.

\bibitem{gallagher2008clothing}
A.~C. Gallagher and T.~Chen, ``Clothing cosegmentation for recognizing
  people,'' in \emph{2008 IEEE CVPR}.\hskip 1em plus 0.5em minus 0.4em\relax
  IEEE, 2008, pp. 1--8.

\bibitem{tsai2016semantic}
Y.-H. Tsai, G.~Zhong, and M.-H. Yang, ``Semantic co-segmentation in videos,''
  in \emph{ECCV}.\hskip 1em plus 0.5em minus 0.4em\relax Springer, 2016, pp.
  760--775.

\bibitem{jerripothula2014automatic}
K.~R. Jerripothula, J.~Cai, F.~Meng, and J.~Yuan, ``Automatic image
  co-segmentation using geometric mean saliency,'' in \emph{IEEE International
  Conference on Image Processing (ICIP)}.\hskip 1em plus 0.5em minus
  0.4em\relax IEEE, 2014, pp. 3277--3281.

\bibitem{jerripothula2015group}
K.~R. Jerripothula, J.~Cai, and J.~Yuan, ``Group saliency propagation for large
  scale and quick image co-segmentation,'' in \emph{IEEE International
  Conference on Image Processing (ICIP)}.\hskip 1em plus 0.5em minus
  0.4em\relax IEEE, 2015, pp. 4639--4643.

\bibitem{meng2012object}
F.~Meng, H.~Li, G.~Liu, and K.~N. Ngan, ``Object co-segmentation based on
  shortest path algorithm and saliency model,'' \emph{IEEE transactions on
  multimedia}, vol.~14, no.~5, pp. 1429--1441, 2012.

\bibitem{rubinstein2013unsupervised}
M.~Rubinstein, A.~Joulin, J.~Kopf, and C.~Liu, ``Unsupervised joint object
  discovery and segmentation in internet images,'' in \emph{IEEE conference on
  computer vision and pattern recognition}, 2013, pp. 1939--1946.

\bibitem{liu2010sift}
C.~Liu, J.~Yuen, and A.~Torralba, ``Sift flow: Dense correspondence across
  scenes and its applications,'' \emph{IEEE transactions on pattern analysis
  and machine intelligence}, vol.~33, no.~5, pp. 978--994, 2010.

\bibitem{tao2017image}
Z.~Tao, H.~Liu, H.~Fu, and Y.~Fu, ``Image cosegmentation via saliency-guided
  constrained clustering with cosine similarity,'' in \emph{AAAI Conference on
  Artificial Intelligence}, vol.~31, no.~1, 2017.

\bibitem{li2018deep}
W.~Li, O.~H. Jafari, and C.~Rother, ``Deep object co-segmentation,'' in
  \emph{Asian Conference on Computer Vision}.\hskip 1em plus 0.5em minus
  0.4em\relax Springer, 2018, pp. 638--653.

\bibitem{chen2018semantic}
H.~Chen, Y.~Huang, and H.~Nakayama, ``Semantic aware attention based deep
  object co-segmentation,'' in \emph{ACCV}.\hskip 1em plus 0.5em minus
  0.4em\relax Springer, 2018, pp. 435--450.

\bibitem{dosovitskiy2015flownet}
A.~Dosovitskiy, P.~Fischer, E.~Ilg, P.~Hausser, C.~Hazirbas, V.~Golkov, P.~Van
  Der~Smagt, D.~Cremers, and T.~Brox, ``Flownet: Learning optical flow with
  convolutional networks,'' in \emph{ICCV}, 2015, pp. 2758--2766.

\bibitem{zhang2020deep}
K.~Zhang, J.~Chen, B.~Liu, and Q.~Liu, ``Deep object co-segmentation via
  spatial-semantic network modulation,'' in \emph{AAAI Conference on Artificial
  Intelligence}, vol.~34, no.~07, 2020, pp. 12\,813--12\,820.

\bibitem{sun2019deep}
K.~Sun, B.~Xiao, D.~Liu, and J.~Wang, ``Deep high-resolution representation
  learning for human pose estimation,'' in \emph{IEEE/CVF Conference on
  Computer Vision and Pattern Recognition}, 2019, pp. 5693--5703.

\bibitem{deselaers2010global}
T.~Deselaers and V.~Ferrari, ``Global and efficient self-similarity for object
  classification and detection,'' in \emph{CVPR}, 2010, pp. 1633--1640.

\bibitem{melekhov2019dgc}
I.~Melekhov, A.~Tiulpin, T.~Sattler, M.~Pollefeys, E.~Rahtu, and J.~Kannala,
  ``Dgc-net: Dense geometric correspondence network,'' in \emph{WACV}, 2019,
  pp. 1034--1042.

\bibitem{9666132}
H.~S. Chhabra and K.~Rao~Jerripothula, ``Comprehensive saliency fusion for
  object co-segmentation,'' in \emph{2021 IEEE International Symposium on
  Multimedia (ISM)}, 2021, pp. 107--110.

\bibitem{deng2009imagenet}
J.~Deng, W.~Dong, R.~Socher, L.-J. Li, K.~Li, and L.~Fei-Fei, ``Imagenet: A
  large-scale hierarchical image database,'' in \emph{2009 IEEE conference on
  computer vision and pattern recognition}.\hskip 1em plus 0.5em minus
  0.4em\relax Ieee, 2009, pp. 248--255.

\bibitem{rousseeuw1987silhouettes}
P.~J. Rousseeuw, ``Silhouettes: a graphical aid to the interpretation and
  validation of cluster analysis,'' \emph{Journal of computational and applied
  mathematics}, vol.~20, pp. 53--65, 1987.

\bibitem{liu2019simple}
J.-J. Liu, Q.~Hou, M.-M. Cheng, J.~Feng, and J.~Jiang, ``A simple pooling-based
  design for real-time salient object detection,'' in \emph{IEEE Conference on
  Computer Vision and Pattern Recognition}, 2019, pp. 3917--3926.

\bibitem{zhao2019egnet}
J.-X. Zhao, J.-J. Liu, D.-P. Fan, Y.~Cao, J.~Yang, and M.-M. Cheng, ``Egnet:
  Edge guidance network for salient object detection,'' in \emph{IEEE/CVF
  International Conference on Computer Vision}, 2019, pp. 8779--8788.

\bibitem{qin2019basnet}
X.~Qin, Z.~Zhang, C.~Huang, C.~Gao, M.~Dehghan, and M.~Jagersand, ``Basnet:
  Boundary-aware salient object detection,'' in \emph{IEEE Conference on
  Computer Vision and Pattern Recognition}, 2019, pp. 7479--7489.

\bibitem{qin2020u2}
X.~Qin, Z.~Zhang, C.~Huang, M.~Dehghan, O.~R. Zaiane, and M.~Jagersand,
  ``U2-net: Going deeper with nested u-structure for salient object
  detection,'' \emph{Pattern Recognition}, vol. 106, p. 107404, 2020.

\bibitem{rother2004grabcut}
C.~Rother, V.~Kolmogorov, and A.~Blake, ``" grabcut" interactive foreground
  extraction using iterated graph cuts,'' \emph{ACM transactions on graphics
  (TOG)}, vol.~23, no.~3, pp. 309--314, 2004.

\bibitem{faktor2013co}
A.~Faktor and M.~Irani, ``Co-segmentation by composition,'' in \emph{IEEE
  international conference on computer vision}, 2013, pp. 1297--1304.

\bibitem{jerripothula2016image}
K.~R. Jerripothula, J.~Cai, and J.~Yuan, ``Image co-segmentation via saliency
  co-fusion,'' \emph{IEEE Transactions on Multimedia}, vol.~18, no.~9, pp.
  1896--1909, 2016.

\bibitem{ren2018mutual}
Y.~Ren, L.~Jiao, S.~Yang, and S.~Wang, ``Mutual learning between saliency and
  similarity: Image cosegmentation via tree structured sparsity and tree graph
  matching,'' \emph{IEEE Transactions on Image Processing}, vol.~27, no.~9, pp.
  4690--4704, 2018.

\bibitem{jerripothula2018quality}
K.~R. Jerripothula, J.~Cai, and J.~Yuan, ``Quality-guided fusion-based
  co-saliency estimation for image co-segmentation and colocalization,''
  \emph{IEEE Transactions on Multimedia}, vol.~20, no.~9, pp. 2466--2477, 2018.

\bibitem{tsai2019image}
C.~C. Tsai, W.~Li, K.~J. Hsu, X.~Qian, and Y.~Y. Lin, ``Image co-saliency
  detection and co-segmentation via progressive joint optimization,''
  \emph{IEEE Transactions on Image Processing}, vol.~28, no.~1, pp. 56--71,
  2019.

\bibitem{tao2019multi}
Z.~Tao, H.~Liu, H.~Fu, and Y.~Fu, ``Multi-view saliency-guided clustering for
  image cosegmentation,'' \emph{IEEE Transactions on Image Processing},
  vol.~28, no.~9, pp. 4634--4645, 2019.

\bibitem{jerripothula2021image}
K.~R. Jerripothula, J.~Cai, J.~Lu, and J.~Yuan, ``Image co-skeletonization via
  co-segmentation,'' \emph{IEEE Transactions on Image Processing}, vol.~30, pp.
  2784--2797, 2021.

\bibitem{quan2016object}
R.~Quan, J.~Han, D.~Zhang, and F.~Nie, ``Object co-segmentation via graph
  optimized-flexible manifold ranking,'' in \emph{CVPR'16}, pp. 687--695.

\bibitem{batra2010icoseg}
D.~Batra, A.~Kowdle, D.~Parikh, J.~Luo, and T.~Chen, ``icoseg: Interactive
  co-segmentation with intelligent scribble guidance,'' in \emph{IEEE Computer
  Vision and Pattern Recognition}, 2010, pp. 3169--3176.

\bibitem{shotton2006textonboost}
J.~Shotton, J.~Winn, C.~Rother, and A.~Criminisi, ``Textonboost: Joint
  appearance, shape and context modeling for multi-class object recognition and
  segmentation,'' in \emph{ECCV}.\hskip 1em plus 0.5em minus 0.4em\relax
  Springer, 2006, pp. 1--15.

\end{thebibliography}

\end{document}